\documentclass[conference]{IEEEtran}
\IEEEoverridecommandlockouts
\usepackage{cite}
\usepackage[T1]{fontenc}
\usepackage{amsmath,amssymb,amsfonts}
\usepackage{amsthm}
\usepackage{graphicx}
\usepackage{booktabs}
\usepackage{textcomp}
\usepackage{xcolor}
\usepackage{url}
\usepackage{tikz}
\usetikzlibrary{positioning,arrows.meta,shapes.geometric,fit,calc,backgrounds}
\usepackage[hidelinks]{hyperref}
\usepackage{xurl} 

\theoremstyle{definition}
\newtheorem{defn}{Definition}

\begin{document}

\title{Mining the Self-Rewriting System: A Vision for Data Mining on AI-Native Infrastructure}

\author{
\IEEEauthorblockN{
Mayank Darbari\IEEEauthorrefmark{1},
Chiman Salavati\IEEEauthorrefmark{2},
Liang Wu\IEEEauthorrefmark{1},
Dawei Wang\IEEEauthorrefmark{1},\\
Jiong Zhang\IEEEauthorrefmark{1},
Kelly Wan\IEEEauthorrefmark{1},
Krishna Chakka\IEEEauthorrefmark{1},
Liangjie Hong\IEEEauthorrefmark{1}
}
\IEEEauthorblockA{
\IEEEauthorrefmark{1}\textit{Nokia}, Sunnyvale, California, USA\\
\{mayank.darbari, liang.wu, dawei.wang, jiong.1.zhang,
kelly.wan, krishna.chakka, liangjie.hong\}@nokia.com
}
\IEEEauthorblockA{
\IEEEauthorrefmark{2}\textit{University of Connecticut},
Storrs, Connecticut, USA\\
chiman.salavati@uconn.edu
}
}

\maketitle

\begin{abstract}
Classical data mining treats data as generated outside the learner. Emerging AI-native infrastructure invalidates this assumption: as agents observe system state, act, update retrieval stores and models, and promote new artifacts into the serving path, their own decisions reshape the future logs, traces, corpora, and evaluation records used for later training and release. We call this new research object a self-rewriting system and the resulting system-induced distribution shift endogenous drift. Our central, falsifiable claim is that cross-surface causes are not
identifiable in general from marginal per-surface monitoring: different
coupled failures can produce the same individual signals while differing
in their joint behavior, requiring joint analysis for reliable attribution. This agenda comprises four foundational research problems: fusing heterogeneous drift signals into calibrated gating and attribution (C1); mining agent traces whose activity alphabet evolves (C2); maintaining knowledge fabrics through online reindexing, faithfulness, and compression decisions (C3); and safely promoting concurrent learning streams through one shared evaluation gate (C4). These remain fragmented across concept drift, performative prediction, process mining, streaming retrieval-augmented generation, and MLOps, without a unified data-mining agenda. We outline public benchmarks and evaluation criteria, using AI-native telecom only as a motivating exemplar with analogues in finance, software CI/CD, and clinical decision support.
\end{abstract}


\section{From Exogenous Data to Endogenous Drift}
\label{sec:intro}

Phone networks, payment systems, software-delivery pipelines, and
clinical decision-support systems are beginning to operate through AI
agents that observe, act, retrieve evidence, update models, and
promote new artifacts back into the serving path. This fundamentally changes the assumptions underlying modern data mining: the system no longer only observes a
data-generating process, it becomes part of it, shaping the streams,
traces, corpora, and evaluation logs from which the next generation
of models and policies is trained. We call this emerging object a
\emph{self-rewriting system}.\footnote{Here self-rewriting means deployed multi-artifact infrastructure loops, not single-model self-revision~\cite{madaan2023selfrefine,yin2024godelagent}.}

\begin{defn}[Self-rewriting system]
\label{def:srs}
A deployed computational system in which model outputs, agent
actions, retrieval updates, tool or policy changes, or promoted
artifacts alter the distribution from which future training,
evaluation, retrieval, or operational data is sampled.
\end{defn}
\begin{defn}[Endogenous drift]
\label{def:endo}
Distributional change induced by a system's own prior decisions,
rather than by an external change in the world alone.
\end{defn}
\begin{defn}[Agent trace]
\label{def:trace}
The structured, parent-linked sequence of \emph{spans} a single
agent session emits, the primary observation signal mined in C2.
\end{defn}
\begin{defn}[Promotion gate]
\label{def:gate}
The shared decision layer that determines whether a newly trained,
distilled, compressed, retrieved, or composed artifact may enter the
serving path.
\end{defn}

The central thesis is that self-rewriting systems render the classical exogenous-data assumption insufficient because the learner becomes part of
the data generator. Classical data mining treats the world as
something to observe; a self-rewriting system observes a world its own
prior actions helped create. The result is not concept drift in one
stream but endogenous drift across coupled artifacts: models,
retrievers, tools, policies, traces, corpora, and promotion logs.

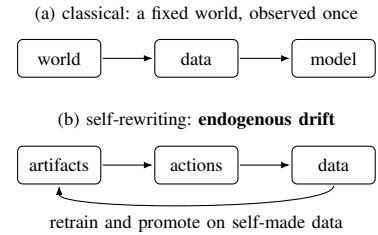
\begin{figure}[t]
\centering
\begin{tikzpicture}[
font=\scriptsize,
b/.style={draw, rounded corners=2pt, align=center,
minimum height=5mm, minimum width=11mm, inner sep=2pt},
lbl/.style={font=\scriptsize, align=center},
arr/.style={-{Latex[length=1.2mm]}, shorten >=0.5pt, shorten <=0.5pt}
]
\node[b] (w)                 {world};
\node[b, right=7mm of w] (d) {data};
\node[b, right=7mm of d] (m) {model};
\draw[arr] (w) -- (d);
\draw[arr] (d) -- (m);
\node[lbl, above=1.0mm of d]
{(a) classical: a fixed world, observed once};

\node[b, below=9mm of w] (m2)     {artifacts};
\node[b, right=7mm of m2] (a2)    {actions};
\node[b, right=7mm of a2] (s2)    {data};
\draw[arr] (m2) -- (a2);
\draw[arr] (a2) -- (s2);
\node[lbl, above=1.0mm of a2]
{(b) self-rewriting: \textbf{endogenous drift}};
\draw[arr]
(s2.south) .. controls +(0,-3.5mm) and +(0,-3.5mm) ..
node[lbl, midway, below=0.2mm]{retrain and promote on self-made data}
(m2.south);
\end{tikzpicture}
\caption{Classical data mining versus a self-rewriting loop.}
\label{fig:endo}
\end{figure}

\textbf{Why now?} The loop is becoming practical because three
ingredients have converged: agents can call external tools through
standard interfaces~\cite{anthropic2024mcp}, adapted models can
be served and hot-swapped on shared
infrastructure~\cite{sheng2024slora,hu2022lora}, and retrieval/evaluation
layers can continuously update the evidence on which agents
act~\cite{edge2024graphrag,es2023ragas}. Early research and industry
activity on AI-native cores~\cite{huawei2025agentic} signal this
direction as an emerging trend, and the pattern is emerging wherever
agents operate critical infrastructure. Individually these components are evaluated as if they observe an external stream; together they form the closed loop in Fig.~\ref{fig:endo}.

\textbf{A concrete failure the loop creates.} A network-operations
agent applies a newly retrieved runbook; the configuration change
shifts telemetry. An anomaly-detection adapter is fine-tuned on the
traces and promoted, and its explanations alter the
citations a root-cause agent emits, which re-enter the corpus that
drives the next retrieval and training slice. When quality later
regresses, the question is not whether one stream drifted but which coupled
artifact to gate: the corpus, tool catalog, adapter, reward model, or
safety policy, since independent detectors may alarm on
several axes yet cannot attribute the cascade the loop itself
created. This is the seed of C1--C4.

\textbf{Why this is a Blue Sky data-mining agenda.} The object of
study is not the agent architecture but the data the loop emits:
heterogeneous drift signals, structured reasoning/action traces,
evolving text/graph/multimodal corpora, and registry-level promotion
logs, that is, streams, sequences, non-stationary graphs, evolving
alphabets, and coupled decision logs, all central objects for data
mining. The Blue Sky step is to treat the loop itself as a new
object: the data stream is no longer external to the mining system,
but partly produced by the same actions, retrieval updates, policy
changes, and promoted artifacts the system later trains, evaluates,
and gates. The opportunity for data mining is to define the public tasks and
metrics for this loop before they are settled ad hoc inside
proprietary stacks; telecom is only a feasibility witness. More precisely, marginal per-surface monitoring is not sufficient in
general for cross-surface attribution: distinct coupled failures can share
the same individual signals but differ in their joint behavior. We treat
this claim as refuted if marginal baselines match joint attribution on the
drift-arena. This single assumption break induces four benchmarkable surfaces:

\begin{itemize}\itemsep1pt\topsep1pt\parskip0pt
\item \textbf{C1. Endogenous drift}: fusing co-evolving
drift signals across corpora, models, tools, policies, and serving
decisions into calibrated gating and attribution decisions.
\item \textbf{C2. Agent-trace mining}: treating structured
reasoning/action traces as primary event streams over evolving tool
and role alphabets.
\item \textbf{C3. Self-evolving knowledge fabrics}: making
reindexing, summary invalidation, citation faithfulness,
provenance tracking, and compression online decisions over
versioned multimodal corpora.
\item \textbf{C4. Closed-loop train-to-serve}: safely promoting
concurrent learning streams through a shared evaluation gate,
registry, observability surface, and safety contract.
\end{itemize}

\section{What Should We Ponder About It?}\label{sec:agenda}

C1--C4 are coupled surfaces, not independent topics: C1 drift signals
can trigger C4 gating; a C4 promotion can change the C3 knowledge
available to agents; that change alters C2 traces, whose outputs feed
future C1 signals. Thus a perturbation on one surface can change the
data-generating process observed by the others, rather than merely
adding correlated features.

\textbf{How this differs from existing agendas.}
The distinction from adjacent areas (concept drift, performative
prediction, process mining, streaming RAG, MLOps) is
\emph{multi-artifact endogeneity}: a coupled loop in which corpora,
tools, policies, adapters, retrieval indices, reward models, serving
decisions, and evaluation gates co-evolve and generate
training and evaluation data. Performative
prediction names model-induced change an \emph{endogenous}
shift, as opposed to environment-driven \emph{exogenous}
drift~\cite{perdomo2020performative,hardt2023performative}; together
with counterfactual / feedback-loop
learning~\cite{bottou2013counterfactual} it is closest in spirit, but
treats a single deployed model rather than an infrastructure loop
whose artifacts co-evolve. We keep this endogenous/exogenous
vocabulary and generalize it from a single-artifact setting to a coupled loop: formally, $s_{t+1}=F(s_t,a_t)$ over an artifact-state vector $s_t$
(corpora, adapters, indices, policies, gates) whose actions $a_t$ and
induced data distribution $\mathcal{D}(s_t)$ both depend on $s_t$;
performative prediction is the single-artifact special case where $s_t$
represents one deployed model. Characterizing $F$ for coupled, heterogeneous
artifacts is an open problem we leave to the agenda. A
related but distinct concern is \emph{model collapse}, the
degradation that follows from training indiscriminately on
model-generated content~\cite{shumailov2024collapse}: that is a
single-model diversity loss, whereas endogenous drift is a
distribution shift across coupled, heterogeneous artifacts that
jointly gate the next release. Each existing agenda isolates one surface (Table~\ref{tab:related}).
The distinctive failure arises when one artifact's action changes
another artifact's future data or evaluation context; without such
cross-artifact feedback, C1--C4 reduce to their existing single-surface
counterparts.

\begin{table*}[t]
\centering
\caption{Comparison with adjacent research areas.}
\label{tab:related}
\scriptsize
\setlength{\tabcolsep}{6pt}
\renewcommand{\arraystretch}{1.0}
\begin{tabular}{@{}p{3.3cm}p{5.9cm}p{7.1cm}@{}}
\toprule
\textbf{Existing area} & \textbf{What it studies} &
\textbf{Why it is not enough for self-rewriting systems}\\
\midrule
\textbf{Concept drift; continual / online learning}~\cite{gama2014drift,bifet2007adwin,webb2016characterizing} &
changing data distributions, and a model adapting to a (possibly
partially observed) stream &
treat the stream as one largely exogenous source; not drift jointly
induced across corpus, tools, policies, adapters, registry, and
serving\\
\addlinespace[2pt]
\textbf{Performative prediction / feedback loops}~\cite{perdomo2020performative,bottou2013counterfactual} &
predictions changing the population they predict &
focuses on one model's decisions; not a multi-artifact
infrastructure loop with heterogeneous, incommensurable signals\\
\addlinespace[2pt]
\textbf{MLOps}~\cite{zaharia2018mlflow} &
the model deployment and release lifecycle &
operationalizes models but defines no mining task over endogenous
traces and coupled promotion interactions\\
\addlinespace[2pt]
\textbf{AIOps}~\cite{dang2019aiops} &
operational telemetry (metrics, logs, alerts) &
mines system signals, not the agent's own structured
reasoning/action trace as the primary data object\\
\addlinespace[2pt]
\textbf{Process mining; business-process drift}~\cite{vanderaalst2016processmining,bose2014processdrift} &
event logs, workflow discovery/conformance, and detecting/localizing
control-flow change in a running process &
handles drift among a roughly fixed activity set; not open-vocabulary
alphabet growth where tools, roles, and skills are added and retired
by the same loop being mined\\
\addlinespace[2pt]
\textbf{Streaming RAG}~\cite{gao2023ragsurvey,edge2024graphrag} &
retrieval over changing corpora &
does not unify citation contracts, summary invalidation, reindex
decisions, multimodal updates, and bounded compression\\
\addlinespace[2pt]
\textbf{Autonomic / self-adaptive systems}~\cite{kephart2003autonomic,delemos2013selfadaptive} &
systems that monitor themselves and adapt (MAPE-K) &
a systems/control framing; lacks benchmarkable data-mining tasks
over traces, drift, corpora, and promotion logs\\
\bottomrule
\end{tabular}
\end{table*}

\begin{table*}[!t]
\centering
\caption{Minimum viable benchmark designs for C1--C4.}
\label{tab:bench}
\scriptsize
\setlength{\tabcolsep}{6pt}
\renewcommand{\arraystretch}{1.0}
\begin{tabular}{@{}p{2.5cm}p{3.4cm}p{3cm}p{2.7cm}p{4.0cm}@{}}
\toprule
\textbf{Benchmark} & \textbf{Data object} &
\textbf{Perturbation / update} & \textbf{Labels / oracle} &
\textbf{Metrics}\\
\midrule
\textbf{Drift-arena} (C1) &
multi-surface drift over corpus, models, tools, policies, serving &
controlled updates with changing cross-surface dependence &
update times, affected surface, root-cause update &
detection delay, attribution F1, joint FPR, gate regret\\
\addlinespace[2pt]
\textbf{Agent-trace} (C2) &
span-level reasoning/action traces &
new tools, role changes, failures, prompt-injection detours &
labeled failures and planted patterns &
per-task F1, latency, tracing overhead, robustness to unseen tools\\
\addlinespace[2pt]
\textbf{Streaming-RAG} (C3) &
versioned multimodal corpus with time-stamped queries &
document updates, stale summaries, query shifts &
answer-validity intervals, gold citations &
time-to-truth, faithfulness, staleness, cost, compression frontier\\
\addlinespace[2pt]
\textbf{Closed-loop} (C4) &
artifact registry with replayable promotion/eval logs &
concurrent promotions, rollbacks, workload shifts &
safe-promotion oracle, non-interference labels &
non-interference, no-skip-eval, rollback time, promotion regret\\
\bottomrule
\end{tabular}
\end{table*}

\subsection{C1. Endogenous drift surfaces}

\emph{Motivation.} Multiple system surfaces may change concurrently,
so the central problem is to identify which change caused a failure
and whether the loop should be gated.

\emph{Problem.} In a self-rewriting loop, drift becomes actionable
when one subsystem's outputs no longer match the next subsystem's
assumptions, and it is endogenous when the mismatch is induced
by the loop's own earlier actions (Def.~\ref{def:endo}). The
difficulty is that drift is not a single stream: the knowledge
surface shifts as specs and runbooks enter the corpus, the model
surface as adapters re-tune and reward models
update~\cite{christiano2017rlhf}, the serving surface as expert
routing shifts mass~\cite{fedus2022switch}, the agent surface as tools
change, and the safety surface as policies are rewritten. The
resulting signals are incommensurable: embedding divergence,
likelihood-ratio shift, reward-function shift, expert-mass shift,
tool-catalog Jaccard distance, and semantic policy difference. C1
asks how to fuse them into one calibrated drift surface that decides
what to attribute, what to gate, and when to keep the loop running.

\emph{Why joint signals can be necessary.} Let binary surface signals
$X,Y$ have two causes: $A$ yields $(0,0)$ or $(1,1)$ equiprobably,
whereas $B$ yields $(0,1)$ or $(1,0)$. Both causes give identical
Bernoulli$(1/2)$ marginals for $X$ and $Y$, so marginal monitoring
cannot distinguish them; successful late fusion must preserve their
cross-surface dependence.

\emph{Benchmark spec.} The drift-arena exposes time-indexed corpus,
model, tool, policy, and serving signals and asks a system to gate and
attribute the root cause, compared with independent, learned, and
causal-graph fusion baselines (Table~\ref{tab:bench}). \emph{Gate regret}
is excess gating loss relative to an oracle gate; the open question is
whether calibration can bound joint FPR when detector dependence is
endogenous.

\subsection{C2. Agent-trace mining}

\emph{Motivation.} Agent sessions produce structured traces of tool
use, retrieval, reasoning, and handoffs, but these traces must be
mined while the available tools, roles, and skills evolve.

\emph{Problem.} Every tool call, model call, sandbox execution,
retrieval step, and human handoff leaves a structured span,
and a session's spans form its agent trace (Def.~\ref{def:trace}). A
span is a tuple of caller, role, tool, argument summary, retrieved
evidence, latency, error class, citation set, parent span, and
outcome, following OpenTelemetry GenAI
conventions~\cite{otelgenai2024} for reasoning-and-acting agent
sessions~\cite{yao2023react}. These traces
carry far more than request/error/duration telemetry because they
expose the agent's intermediate decisions. The mining tasks include
detecting capability gaps, slow tool paths, redundant retrievals,
citation failures, prompt-injection
detours~\cite{greshake2023injection,perez2022injection}, role mismatch,
and reusable span subsequences that can become new
skills~\cite{wang2023voyager}.

\emph{What is missing.} Prior work spans process mining, agent telemetry,
agent benchmarks, retrieval evaluation, and LLM-as-judge
evaluation~\cite{vanderaalst2016processmining,otelgenai2024,
liu2024agentbench,zhou2024webarena,jimenez2024swebench,
xie2024osworld,yao2024taubench,wang2023voyager,es2023ragas,
zheng2023judging}. However, these lines of work do not yet formulate
a public data-mining task for agent traces whose event alphabet
evolves during deployment. Business-process
drift~\cite{bose2014processdrift} studies control-flow change over a
largely fixed activity set; in self-rewriting systems, by contrast,
the alphabet is endogenous: tools, roles, and skills may be added,
modified, or retired by the same loop that generates the trace data.

\emph{Benchmark spec.} An agent-trace benchmark provides sessions of
spans and poses the mining tasks above over an evolving tool and
role alphabet, scored for robustness to tool additions/removals and
generalization to unseen tools (Table~\ref{tab:bench}). The key open
question is how to define frequent or useful patterns when the span
alphabet grows, shrinks, and changes meaning during deployment.

\subsection{C3. Self-evolving knowledge fabrics}

\emph{Motivation.} In self-rewriting systems, the knowledge sources
used by an agent evolve over time, so answer validity, citation
faithfulness, and summary consistency must be maintained under
continuous corpus updates.

\emph{Problem.} The hard part of a self-rewriting system's knowledge
fabric is not retrieval alone but the stream of \emph{online
decisions} required as the corpus changes underneath the agent: when
to reindex, when to invalidate a summary, when a citation is no longer valid with respect to the current corpus, how to keep faithfulness over versioned corpora, and how to
compress context under a bounded budget. Batch faithfulness
scoring~\cite{es2023ragas} hides these because it never asks when a
corpus update becomes visible to future answers. C3 treats citation
as an online contract, summary maintenance as a streaming graph
problem, reindexing as a drift-triggered decision, multimodal updates
as first-class corpus events (vision-language models such as
LLaVA~\cite{liu2023llava} make image and text jointly indexable), and
compression as a budgeted fidelity
trade-off~\cite{edge2024graphrag,sarthi2024raptor,jiang2023llmlingua}.

\emph{Beyond streaming RAG.} Graph-augmented retrieval~\cite{edge2024graphrag},
hierarchical summarization~\cite{sarthi2024raptor}, and faithfulness
evaluation~\cite{es2023ragas} largely address retrieval quality and corpus
maintenance~\cite{gao2023ragsurvey}. The key difference here is endogeneity:
answers, citations, summaries, and promoted artifacts can re-enter the corpus,
so retrieval decisions alter future evidence. C3 studies this
feedback loop across versioned citation, summary invalidation, multimodal
updates, and compression. We measure \emph{time-to-truth}, the lag from an
update to the first answer reflecting it, and the
\emph{compression-faithfulness frontier}, the best faithfulness at each
compression ratio under a fixed retrieval budget.

\emph{Benchmark spec.} A streaming-RAG benchmark provides a
versioned, time-stamped multimodal corpus with known update events,
queries, answer-validity intervals, and a retrieval budget, scored
by time-to-truth and the compression-faithfulness
frontier among other metrics (Table~\ref{tab:bench}). The key open
question is whether freshness and faithfulness can be optimized
together, or whether a fixed inference budget imposes a fundamental
trade-off.

\subsection{C4. Closed-loop train-to-serve}

\emph{Motivation.} Multiple artifacts may be updated concurrently and
promoted through one shared gate, so release decisions must account
for cross-stream interactions.

\emph{Problem.} The data-mining challenge is not how to optimize
one training stream; it is how to decide which of many interacting
streams may safely enter a shared serving path. Concurrent updates to adapters, preference models, distilled models,
routing policies, or skill
libraries~\cite{rafailov2023dpo,christiano2017rlhf,hinton2015distillation,hu2022lora,fedus2022switch,jiang2023llmlingua,wang2023voyager}
each follow a path from production traces to a trained artifact,
evaluation log, and serving decision. These streams share one
registry, evaluation surface, and safety contract behind a single
promotion gate (Def.~\ref{def:gate}), so a promotion in one stream
can change the evidence, workload, or evaluation context seen by
another. C4 asks how to mine promotion logs and evaluation traces to
decide when to promote, delay, isolate, retest, or roll back an
artifact.

\emph{What is missing.} MLOps~\cite{zaharia2018mlflow}
and streaming ML~\cite{gama2014drift,webb2016characterizing} take a
single-artifact or single-policy view; C4 instead studies heterogeneous
streams sharing one gate under properties such as no-skip-eval,
promotion provenance, and multi-stream non-interference
(Def.~\ref{def:nonint}). In an internal deployment (details withheld),
tightening one stream's gate from 50 to 200 targets reduced judge-graded
quality from 2.78 to 2.60 on 500 held-out prompts, recovering to 3.15
with LLM-judge gating. A public counterpart can inject concurrent
promotions and replay each with and without the other stream's update
to measure cross-stream interference.

\begin{defn}[Multi-Stream Non-Interference]
\label{def:nonint}
For concurrent streams $s_i, s_j$ sharing registry $\mathcal{R}$,
with promotion events $p_i, p_j$ and per-stream evaluator
$\mathrm{eval}_j$:
\[
\bigl|\mathrm{eval}_j(p_j \mid p_i\!\to\!\mathcal{R})
      -\mathrm{eval}_j(p_j)\bigr| < \epsilon_j .
\]
\emph{Detection.} Re-run $\mathrm{eval}_j$ against $p_j$ after $p_i$
is registered; the observed gap is the empirical left-hand side,
surfaced by the registry's promotion-replay log.
\end{defn}

\emph{Benchmark spec.} A closed-loop benchmark provides replayable
promotion/evaluation logs and asks a system to promote, delay, rollback,
isolate, or retest under a shared safety contract (Table~\ref{tab:bench}).
\emph{No-skip-eval} is the fraction of promotions completing all required
evaluations (target one); \emph{promotion regret} is excess deployment
loss relative to the safe-promotion oracle. The open question is whether
safe concurrency can bound cross-stream interference.

\section{What Is the Blue Sky Idea?}
\label{sec:exist}

The Blue Sky idea is to treat the self-rewriting system as a primary
unit of study and to turn C1--C4 into a shared, open benchmark
agenda: a drift-arena (C1), an agent-trace bench (C2), a
streaming-RAG bench (C3), and a closed-loop promotion bench (C4),
each turning a hidden assumption into a measurable task. The bet is
that if the community builds these in the open, the core problems of
self-rewriting systems will be defined by public tasks rather than
private production conventions.

\textbf{A public benchmark plan.} Because production traces are often
proprietary, public benchmarks can use public corpora and synthetic
generators with controlled perturbations, while internal trajectories only
inform the schema and failure modes. Generators should vary coupled updates,
evolving alphabets, non-IID/federated clients, and adversarial events such as poisoning or prompt injection. Recording the affected artifact, root-cause
axis, and counterfactual no-promotion baseline provides ground truth for
attribution and non-interference.

\section{What Will Success Look Like?}
\label{sec:success}

Success progresses from public C1--C4 benchmarks and baselines in the
near term to shared metrics and theory for endogenous drift, trace
mining, knowledge maintenance, and safe promotion. Long term,
self-rewriting systems should report reproducible cross-surface
evidence across domains, not model quality alone.

\section{Implications for Data Mining}
\label{sec:care}

Self-rewriting loops recur beyond telecom: finance agents reshape
future monitoring, CI/CD agents alter later tests and training logs,
and clinical recommendations affect future notes and retrieval.
Across domains, system outputs become inputs to the next loop.

C1--C4 are new forms of objects the data-mining community already
studies: streams, traces, graphs, and coupled decision logs. The
community that built concept drift, process
mining, streaming algorithms, and retrieval evaluation is best
positioned to define these public tasks and metrics before they settle inside private
stacks: self-rewriting systems should become a first-class object of
data mining, and C1--C4 provide the initial benchmark map. When cross-surface causes are not identifiable from marginal signals,
systems deployed today may be evaluated with tools blind to important
loop-induced failure modes.

{\footnotesize
\bibliographystyle{IEEEtran}
\bibliography{refs}}

@article{huawei2025agentic,
  author    = {Li Xu and Weisen Shi and Hang Zhang and Cheng-Hui Peng and
               Shizhou Wu and Wen Tong},
  title     = {The Agentic-{AI} Core: An {AI}-Empowered, Mission-Oriented
               Core Network for Next-Generation Mobile Telecommunications},
  journal   = {Engineering},
  year      = {2025},
  note      = {In press. \url{https://doi.org/10.1016/j.eng.2025.06.027}}
}

@inproceedings{yao2023react,
  author    = {Shunyu Yao and Jeffrey Zhao and Dian Yu and Nan Du and
               Izhak Shafran and Karthik Narasimhan and Yuan Cao},
  title     = {{ReAct}: Synergizing Reasoning and Acting in Language Models},
  booktitle = {Proc. ICLR},
  year      = {2023},
  note      = {\url{https://arxiv.org/abs/2210.03629}}
}

@inproceedings{greshake2023injection,
  author    = {Kai Greshake and Sahar Abdelnabi and Shailesh Mishra and
               Christoph Endres and Thorsten Holz and Mario Fritz},
  title     = {Not What You've Signed Up For: Compromising Real-World
               {LLM}-Integrated Applications with Indirect Prompt Injection},
  booktitle = {Proc. 16th ACM Workshop on Artificial Intelligence and
               Security (AISec)},
  year      = {2023},
  pages     = {79--90},
  note      = {\url{https://arxiv.org/abs/2302.12173}}
}

@inproceedings{sheng2024slora,
  author    = {Ying Sheng and Shiyi Cao and Dacheng Li and Coleman Hooper
               and Nicholas Lee and Shuo Yang and Christopher Chou and
               Banghua Zhu and Lianmin Zheng and Kurt Keutzer and
               Joseph E. Gonzalez and Ion Stoica},
  title     = {{S-LoRA}: Serving Thousands of Concurrent {LoRA} Adapters},
  booktitle = {Proc. Conf. Machine Learning and Systems (MLSys)},
  year      = {2024},
  note      = {\url{https://arxiv.org/abs/2311.03285}}
}

@misc{anthropic2024mcp,
  author       = {{Anthropic}},
  title        = {Introducing the Model Context Protocol},
  year         = {2024},
  howpublished = {Anthropic engineering blog},
  note         = {\url{https://www.anthropic.com/news/model-context-protocol}}
}

@inproceedings{sarthi2024raptor,
  author    = {Parth Sarthi and Salman Abdullah and Aditi Tuli and
               Shubh Khanna and Anna Goldie and Christopher D. Manning},
  title     = {{RAPTOR}: Recursive Abstractive Processing for
               Tree-Organized Retrieval},
  booktitle = {Proc. 12th Int. Conf. Learning Representations (ICLR)},
  year      = {2024},
  note      = {\url{https://arxiv.org/abs/2401.18059}}
}

@inproceedings{hu2022lora,
  author    = {Edward J. Hu and others},
  title     = {{LoRA}: Low-Rank Adaptation of Large Language Models},
  booktitle = {Proc. ICLR},
  year      = {2022},
  note      = {\url{https://arxiv.org/abs/2106.09685}}
}

@article{zaharia2018mlflow,
  author    = {Matei Zaharia and others},
  title     = {Accelerating the Machine Learning Lifecycle with
               {MLflow}},
  journal   = {IEEE Data Eng. Bull.},
  volume    = {41},
  number    = {4},
  pages     = {39--45},
  year      = {2018},
  note      = {\url{http://sites.computer.org/debull/A18dec/p39.pdf}}
}

@article{edge2024graphrag,
  author    = {Darren Edge and Ha Trinh and Newman Cheng and Joshua Bradley
               and Alex Chao and Apurva Mody and Steven Truitt and
               Jonathan Larson},
  title     = {From Local to Global: A {Graph RAG} Approach to
               Query-Focused Summarization},
  journal   = {arXiv preprint},
  year      = {2024},
  note      = {\url{https://arxiv.org/abs/2404.16130}}
}

@inproceedings{es2023ragas,
  author    = {Shahul Es and Jithin James and Luis Espinosa-Anke and
               Steven Schockaert},
  title     = {{RAGAS}: Automated Evaluation of Retrieval Augmented
               Generation},
  booktitle = {Proc. 18th Conf. European Chapter of the ACL (EACL):
               System Demonstrations},
  year      = {2024},
  note      = {\url{https://arxiv.org/abs/2309.15217}}
}

@inproceedings{zheng2023judging,
  author    = {Lianmin Zheng and Wei-Lin Chiang and Ying Sheng and
               Siyuan Zhuang and Zhanghao Wu and Yonghao Zhuang and
               Zi Lin and Zhuohan Li and Dacheng Li and Eric P. Xing and
               Hao Zhang and Joseph E. Gonzalez and Ion Stoica},
  title     = {Judging {LLM}-as-a-Judge with {MT-Bench} and {Chatbot Arena}},
  booktitle = {Proc. NeurIPS Datasets and Benchmarks Track},
  year      = {2023},
  note      = {\url{https://arxiv.org/abs/2306.05685}}
}

@article{gama2014drift,
  author    = {Jo{\~a}o Gama and Indr{\.e} {\v{Z}}liobait{\.e} and
               Albert Bifet and Mykola Pechenizkiy and
               Abdelhamid Bouchachia},
  title     = {A Survey on Concept Drift Adaptation},
  journal   = {ACM Computing Surveys},
  volume    = {46},
  number    = {4},
  year      = {2014},
  note      = {\url{https://doi.org/10.1145/2523813}}
}

@inproceedings{bifet2007adwin,
  author    = {Albert Bifet and Ricard Gavald{\`a}},
  title     = {Learning from Time-Changing Data with Adaptive Windowing},
  booktitle = {Proc. SIAM Int. Conf. Data Mining (SDM)},
  year      = {2007},
  pages     = {443--448},
  note      = {\url{https://doi.org/10.1137/1.9781611972771.42}}
}

@article{webb2016characterizing,
  author    = {Geoffrey I. Webb and Roy Hyde and Hong Cao and
               Hai Long Nguyen and Fran{\c{c}}ois Petitjean},
  title     = {Characterizing Concept Drift},
  journal   = {Data Mining and Knowledge Discovery},
  volume    = {30},
  number    = {4},
  pages     = {964--994},
  year      = {2016},
  note      = {\url{https://arxiv.org/abs/1511.03816}}
}

@book{vanderaalst2016processmining,
  author    = {Wil M.~P. van der Aalst},
  title     = {Process Mining: Data Science in Action},
  edition   = {2nd},
  publisher = {Springer},
  year      = {2016},
  note      = {\url{https://doi.org/10.1007/978-3-662-49851-4}}
}

@misc{otelgenai2024,
  author       = {{OpenTelemetry Authors}},
  title        = {Semantic Conventions for {Generative AI} Systems},
  year         = {2024},
  howpublished = {OpenTelemetry specification},
  note         = {\url{https://opentelemetry.io/docs/specs/semconv/gen-ai/}}
}

@inproceedings{liu2024agentbench,
  author    = {Xiao Liu and Hao Yu and Hanchen Zhang and Yifan Xu and
               Xuanyu Lei and Hanyu Lai and Yu Gu and Hangliang Ding and
               Kaiwen Men and Kejuan Yang and Shudan Zhang and Xiang Deng
               and Aohan Zeng and Zhengxiao Du and Chenhui Zhang and
               Sheng Shen and Tianjun Zhang and Yu Su and Huan Sun and
               Minlie Huang and Yuxiao Dong and Jie Tang},
  title     = {{AgentBench}: Evaluating {LLMs} as Agents},
  booktitle = {Proc. Int. Conf. Learning Representations (ICLR)},
  year      = {2024},
  note      = {\url{https://arxiv.org/abs/2308.03688}}
}

@inproceedings{perez2022injection,
  author    = {F{\'a}bio Perez and Ian Ribeiro},
  title     = {Ignore Previous Prompt: Attack Techniques for Language
               Models},
  booktitle = {NeurIPS Workshop on Machine Learning Safety},
  year      = {2022},
  note      = {\url{https://arxiv.org/abs/2211.09527}}
}

@article{gao2023ragsurvey,
  author    = {Yunfan Gao and Yun Xiong and Xinyu Gao and Kangxiang Jia
               and Jinliu Pan and Yuxi Bi and Yi Dai and Jiawei Sun and
               Qianyu Guo and Meng Wang and Haofen Wang},
  title     = {Retrieval-Augmented Generation for Large Language Models:
               A Survey},
  journal   = {arXiv preprint},
  year      = {2023},
  note      = {\url{https://arxiv.org/abs/2312.10997}}
}

@inproceedings{rafailov2023dpo,
  author    = {Rafael Rafailov and Archit Sharma and Eric Mitchell and
               Stefano Ermon and Christopher D. Manning and Chelsea Finn},
  title     = {Direct Preference Optimization: Your Language Model is
               Secretly a Reward Model},
  booktitle = {Proc. NeurIPS},
  year      = {2023},
  note      = {\url{https://arxiv.org/abs/2305.18290}}
}

@inproceedings{christiano2017rlhf,
  author    = {Paul F. Christiano and Jan Leike and Tom B. Brown and
               Miljan Martic and Shane Legg and Dario Amodei},
  title     = {Deep Reinforcement Learning from Human Preferences},
  booktitle = {Proc. NeurIPS},
  year      = {2017},
  note      = {\url{https://arxiv.org/abs/1706.03741}}
}

@inproceedings{hinton2015distillation,
  author    = {Geoffrey Hinton and Oriol Vinyals and Jeff Dean},
  title     = {Distilling the Knowledge in a Neural Network},
  booktitle = {NeurIPS Deep Learning and Representation Learning Workshop},
  year      = {2015},
  note      = {\url{https://arxiv.org/abs/1503.02531}}
}

@article{fedus2022switch,
  author    = {William Fedus and Barret Zoph and Noam Shazeer},
  title     = {Switch Transformers: Scaling to Trillion Parameter Models
               with Simple and Efficient Sparsity},
  journal   = {Journal of Machine Learning Research},
  volume    = {23},
  number    = {120},
  pages     = {1--39},
  year      = {2022},
  note      = {\url{https://arxiv.org/abs/2101.03961}}
}

@inproceedings{jiang2023llmlingua,
  author    = {Huiqiang Jiang and Qianhui Wu and Chin-Yew Lin and Yuqing Yang
               and Lili Qiu},
  title     = {{LLMLingua}: Compressing Prompts for Accelerated Inference of
               Large Language Models},
  booktitle = {Proc. EMNLP},
  year      = {2023},
  note      = {\url{https://arxiv.org/abs/2310.05736}}
}

@inproceedings{liu2023llava,
  author    = {Haotian Liu and Chunyuan Li and Qingyang Wu and Yong Jae Lee},
  title     = {Visual Instruction Tuning},
  booktitle = {Proc. NeurIPS},
  year      = {2023},
  note      = {\url{https://arxiv.org/abs/2304.08485}}
}

@article{wang2023voyager,
  author    = {Guanzhi Wang and Yuqi Xie and Yunfan Jiang and Ajay Mandlekar
               and Chaowei Xiao and Yuke Zhu and Linxi Fan and Anima Anandkumar},
  title     = {Voyager: An Open-Ended Embodied Agent with Large Language
               Models},
  journal   = {arXiv preprint},
  year      = {2023},
  note      = {\url{https://arxiv.org/abs/2305.16291}}
}

@inproceedings{zhou2024webarena,
  author    = {Shuyan Zhou and Frank F. Xu and Hao Zhu and Xuhui Zhou and
               Robert Lo and Abishek Sridhar and Xianyi Cheng and Tianyue
               Ou and Yonatan Bisk and Daniel Fried and Uri Alon and
               Graham Neubig},
  title     = {{WebArena}: A Realistic Web Environment for Building
               Autonomous Agents},
  booktitle = {Proc. Int. Conf. Learning Representations (ICLR)},
  year      = {2024},
  note      = {\url{https://arxiv.org/abs/2307.13854}}
}

@inproceedings{jimenez2024swebench,
  author    = {Carlos E. Jimenez and John Yang and Alexander Wettig and
               Shunyu Yao and Kexin Pei and Ofir Press and Karthik
               Narasimhan},
  title     = {{SWE-bench}: Can Language Models Resolve Real-World GitHub
               Issues?},
  booktitle = {Proc. Int. Conf. Learning Representations (ICLR)},
  year      = {2024},
  note      = {\url{https://arxiv.org/abs/2310.06770}}
}

@inproceedings{xie2024osworld,
  author    = {Tianbao Xie and Danyang Zhang and Jixuan Chen and Xiaochuan
               Li and Siheng Zhao and Ruisheng Cao and Toh Jing Hua and
               Zhoujun Cheng and Dongchan Shin and Fangyu Lei and Yitao
               Liu and Yiheng Xu and Shuyan Zhou and Silvio Savarese and
               Caiming Xiong and Victor Zhong and Tao Yu},
  title     = {{OSWorld}: Benchmarking Multimodal Agents for Open-Ended
               Tasks in Real Computer Environments},
  booktitle = {Proc. NeurIPS Datasets and Benchmarks Track},
  year      = {2024},
  note      = {\url{https://arxiv.org/abs/2404.07972}}
}

@inproceedings{yao2024taubench,
  author    = {Shunyu Yao and Noah Shinn and Pedram Razavi and Karthik R.
               Narasimhan},
  title     = {{$\tau$-Bench}: A Benchmark for Tool-Agent-User Interaction
               in Real-World Domains},
  booktitle = {Proc. Int. Conf. Learning Representations (ICLR)},
  year      = {2025},
  note      = {\url{https://arxiv.org/abs/2406.12045}}
}

@inproceedings{perdomo2020performative,
  author    = {Juan C. Perdomo and Tijana Zrnic and
               Celestine Mendler-D{\"u}nner and Moritz Hardt},
  title     = {Performative Prediction},
  booktitle = {Proc. 37th Int. Conf. Machine Learning (ICML)},
  pages     = {7599--7609},
  year      = {2020},
  note      = {\url{https://arxiv.org/abs/2002.06673}}
}

@article{bottou2013counterfactual,
  author    = {L{\'e}on Bottou and Jonas Peters and
               Joaquin Qui{\~n}onero-Candela and Denis X. Charles and
               D. Max Chickering and Elon Portugaly and Dipankar Ray and
               Patrice Simard and Ed Snelson},
  title     = {Counterfactual Reasoning and Learning Systems: The Example
               of Computational Advertising},
  journal   = {Journal of Machine Learning Research},
  volume    = {14},
  number    = {1},
  pages     = {3207--3260},
  year      = {2013},
  note      = {\url{https://www.jmlr.org/papers/v14/bottou13a.html}}
}

@article{kephart2003autonomic,
  author    = {Jeffrey O. Kephart and David M. Chess},
  title     = {The Vision of Autonomic Computing},
  journal   = {Computer},
  volume    = {36},
  number    = {1},
  pages     = {41--50},
  year      = {2003},
  note      = {\url{https://doi.org/10.1109/MC.2003.1160055}}
}

@incollection{delemos2013selfadaptive,
  author    = {Rog{\'e}rio de Lemos and Holger Giese and
               Hausi A. M{\"u}ller and Mary Shaw and others},
  title     = {Software Engineering for Self-Adaptive Systems: A Second
               Research Roadmap},
  booktitle = {Software Engineering for Self-Adaptive Systems II},
  series    = {Lecture Notes in Computer Science},
  volume    = {7475},
  publisher = {Springer},
  pages     = {1--32},
  year      = {2013},
  note      = {\url{https://doi.org/10.1007/978-3-642-35813-5_1}}
}

@inproceedings{dang2019aiops,
  author    = {Yingnong Dang and Qingwei Lin and Peng Zhang},
  title     = {{AIOps}: Real-World Challenges and Research Innovations},
  booktitle = {Proc. IEEE/ACM 41st Int. Conf. Software Engineering:
               Companion (ICSE-Companion)},
  pages     = {4--5},
  year      = {2019},
  note      = {\url{https://doi.org/10.1109/ICSE-Companion.2019.00023}}
}

@article{shumailov2024collapse,
  author    = {Ilia Shumailov and Zakhar Shumaylov and Yiren Zhao and
               Nicolas Papernot and Ross Anderson and Yarin Gal},
  title     = {{AI} Models Collapse When Trained on Recursively Generated
               Data},
  journal   = {Nature},
  volume    = {631},
  number    = {8022},
  pages     = {755--759},
  year      = {2024},
  note      = {\url{https://doi.org/10.1038/s41586-024-07566-y}}
}

@article{bose2014processdrift,
  author    = {R.~P. Jagadeesh Chandra Bose and Wil M.~P. van der Aalst and
               Indr{\.e} {\v{Z}}liobait{\.e} and Mykola Pechenizkiy},
  title     = {Dealing With Concept Drifts in Process Mining},
  journal   = {IEEE Transactions on Neural Networks and Learning Systems},
  volume    = {25},
  number    = {1},
  pages     = {154--171},
  year      = {2014},
  note      = {\url{https://doi.org/10.1109/TNNLS.2013.2278313}}
}

@article{hardt2023performative,
  author    = {Moritz Hardt and Celestine Mendler-D{\"u}nner},
  title     = {Performative Prediction: Past and Future},
  journal   = {arXiv preprint},
  year      = {2023},
  note      = {\url{https://arxiv.org/abs/2310.16608}}
}

@article{yin2024godelagent,
  author    = {Xunjian Yin and Xinyi Wang and Liangming Pan and Li Lin and
               Xiaojun Wan and William Yang Wang},
  title     = {{G{\"o}del Agent}: A Self-Referential Agent Framework for
               Recursive Self-Improvement},
  journal   = {arXiv preprint},
  year      = {2024},
  note      = {\url{https://arxiv.org/abs/2410.04444}}
}

@inproceedings{madaan2023selfrefine,
  author    = {Aman Madaan and Niket Tandon and Prakhar Gupta and
               Skyler Hallinan and Luyu Gao and Sarah Wiegreffe and
               Uri Alon and Nouha Dziri and Shrimai Prabhumoye and
               Yiming Yang and Shashank Gupta and
               Bodhisattwa Prasad Majumder and Katherine Hermann and
               Sean Welleck and Amir Yazdanbakhsh and Peter Clark},
  title     = {{Self-Refine}: Iterative Refinement with Self-Feedback},
  booktitle = {Proc. NeurIPS},
  year      = {2023},
  note      = {\url{https://arxiv.org/abs/2303.17651}}
}

\end{document}